\documentclass[letterpaper, 10 pt, conference]{ieeeconf}

\IEEEoverridecommandlockouts
\usepackage{cite}
\usepackage[hidelinks]{hyperref}       
\usepackage{graphicx}
\usepackage{amsmath}
\usepackage{amsfonts}
\usepackage{algorithm}
\usepackage{algpseudocode}
\usepackage{booktabs, multirow} 
\usepackage{soul}
\usepackage[table]{xcolor} 
\usepackage{changepage,threeparttable} 

\DeclareMathOperator*{\argmax}{arg\,max}

\title{Dynamics and Domain Randomized Gait Modulation with Bezier Curves for Sim-to-Real Legged Locomotion}
\author{
    Maurice Rahme$^1$, Ian Abraham$^2$, Matthew L. Elwin$^1$, Todd D. Murphey$^1$%
    \thanks{*This work is partly supported by the Northwestern Robotics program.
            In addition, this material is based upon work supported by the National Science Foundation under Grants CNS 1837515.
            Any opinions, findings and conclusions or recommendations expressed in this material are those of the authors and
            do not necessarily reflect the views of the aforementioned institutions.
    }
    \thanks{$^1$ Department of Mechanical Engineering, Northwestern University,
            Evanston, IL 60208}%
 \thanks{$^2$ Robotics Institute at Carnegie Mellon University, Pittsburgh, PA 15213}%
\thanks{Corresponding Email: {\tt\footnotesize mauricerahme2020@u.northwestern.edu},
        {\tt\small ia@andrew.cmu.edu} } %
    \thanks{Multimedia can be found at \url{https://sites.google.com/view/d2gmbc/home}.}
}

\begin{document}
\maketitle
\thispagestyle{empty}
\pagestyle{empty}

\begin{abstract}
  We present a sim-to-real framework that uses dynamics and domain randomized offline reinforcement learning to enhance
  open-loop gaits for legged robots, allowing them to traverse uneven terrain without sensing foot impacts.
  Our approach, D$^2$-Randomized Gait Modulation with Bezier Curves (D$^2$-GMBC), uses augmented random search with randomized dynamics and terrain to train,
  in simulation, a policy that modifies the parameters and output of an
  open-loop Bezier curve gait generator for quadrupedal robots.
  The policy, using only inertial measurements, enables the robot to traverse unknown rough terrain, even when the robot's physical parameters do
  not match the open-loop model.

  We compare the resulting policy to hand-tuned Bezier Curve gaits and to policies trained without randomization, both
  in simulation and on a real quadrupedal robot. With D$^2$-GMBC, across a variety of experiments on unobserved and unknown uneven terrain,
  the robot walks significantly farther than with either hand-tuned gaits or gaits learned without domain randomization.
  Additionally, using D$^2$-GMBC, the robot can walk laterally and rotate while on the rough terrain, even though it was trained only for forward walking.
\end{abstract}

\section{Introduction}\label{sec:intro}

Legged robots can reach areas inaccessible to their wheeled
counterparts by stepping over obstacles and shifting their center of mass to avoid falling.
However, rough terrain is highly variable and difficult to model: to traverse it, robots
must either explicitly sense and adjust or, as in this paper, rely on gaits that are insensitive to ground variation.

\begin{figure} 
  \centering
  \includegraphics[width=0.5\textwidth]{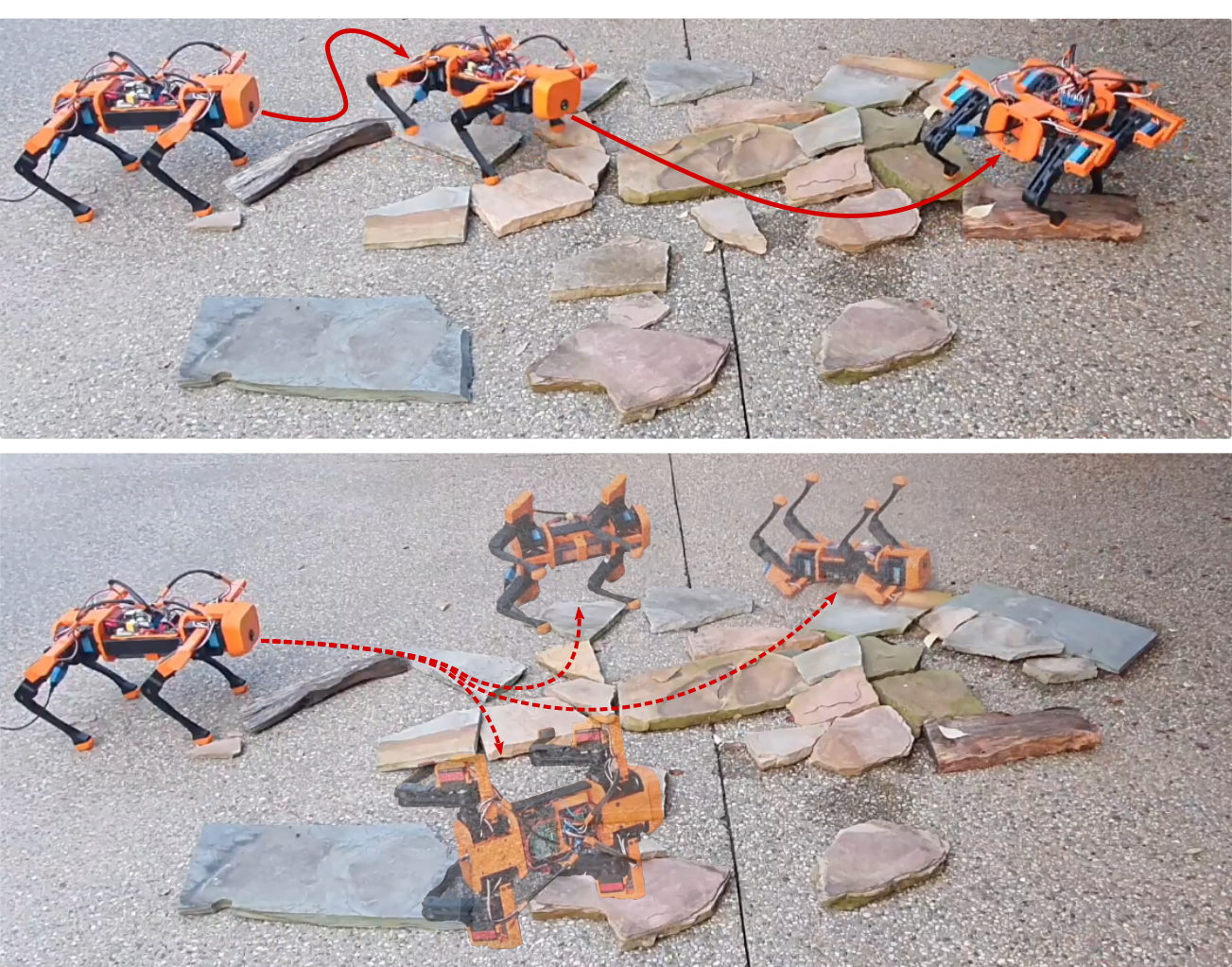}
  \caption{
    (TOP) Example of sim-to-real transfer of D$^2$-GMBC evaluated
    on a real robot in difficult terrain. (BOTTOM) A legged robot using an open-loop
    hand-tuned Bezier curve gait generator failing to traverse the terrain due to
    brittleness of open-loop gaits.
  }
  \label{Intro}
\end{figure}

We focus on quadrupedal locomotion in unobserved and unmodeled rough terrain. We extend the hybrid learning approach
of~\cite{iscen2018policies}, which uses a learned policy in conjunction with an open-loop dynamics model to control a system.
The model provides baseline behavior while the policy uses sensor feedback to adapt the model and improve the control.
Specifically, we model leg trajectories using Bezier curve gaits from~\cite{cheetah2014} and augment
the underlying gait parameters using a policy trained with reinforcement learning (RL)
in simulation. By using dynamics randomization (\cite{heess2017emergence,tan2018sim}) while training,
the resulting policies produce gaits that, in our experiments and without modification, adapt to unmodeled terrain and dynamics.

Although many RL techniques can be used within our framework,
we use augmented random search (ARS)~\cite{mania2018simple} for optimizing policy parameters because
the resulting policies are linear, enabling efficient implementation on devices with limited computational power.

During training, we provide variable external steering commands to point the robot forward, resulting
in a policy that handles steering inputs (from a user or motion planner) without additional training.

The primary contribution of this paper is the Dynamics and Domain Randomized Gait Modulation with Bezier Curves (D$^2$-GMBC) framework,
a hybrid approach to legged locomotion combining reinforcement learning and open-loop gaits. With D$^2$GMBC:
\begin{enumerate}
\item Policies trained in simulation can be directly deployed on real robots to enable them to walk over rough terrain.
      We test this sim-to-real transfer experimentally.
\item Walking on rough terrain requires only inertial measurements: no foot contact or terrain sensing required.
\item We compare D$^2$-GMBC to a similar method without randomized terrain, which serves as a proxy for other
  methods that randomize dynamics but train on flat ground (e.g.,~\cite{peng2018sim}). D$^2$-GMBC outperforms the non-randomized terrain
  method in simulated tests over rough terrain, despite performing worse in training.
\end{enumerate}

We also provide plans for the OpenQuadraped (the robot used in our experiments)~\cite{openquadruped2020}, an open-source simulation environment~\cite{spotminimini2020github}, and additional media at \url{https://sites.google.com/view/d2gmbc/home}.


\section{Related Work}\label{sec:related}
    Methods for finding suitable gaits for legged locomotion can be
    broadly classified into methods that require modeling and tuning~\cite{Raibert2008,Tedrake2011,
      pratt2001virtual,todd2013walking,shigley1960mechanics,cheetah2014,Ruina2014,saranli2001rhex,Gehring2013}, evolutionary approaches~\cite{lipson2006evolutionary,BongardLipson2006Science,CurrieEtAl2008, ChernovaVeloso2004},
    and learning approaches~\cite{rieffel2010morphological, iscen2018policies,heess2017emergence,tan2018sim,kohl2004policy}.

  We focus on three classes of learning approaches here, all of which are trained in simulation and then (mostly)  deployed
  (sometimes with modification) on a robot: modulating gait generators, sim-to-real, and imitation learning.

    \textbf{Modulating Gait Generators:} This approach extends existing model-based locomotive gaits
    by augmenting them via learning. The model-based gait provides a baseline
    which is improved by a learned residual function. The work of~\cite{iscen2018policies} augments sine trajectory gait generators with policies
    learned in simulation from reinforcement learning. These policies are transferred to real robots to improve forward
    velocity movement on flat ground. Similarly, \cite{kohl2004policy} uses reinforcement
    learning on robots to augment elliptical gaits and increase forward speed with three
    hours of training on a real robot.

    We extend these ideas to a broader class of gait generators (i.e., Bezier curves) and
    situations (rough terrain, lateral and rotational motion). Thus, our approach extends the robot's capabilities by enabling
    a human operator or motion planner to provide directional commands.

    \textbf{Sim-to-real:} This legged locomotion approach focuses on
    learning a policy in simulation and transferring it to a real robot without requiring a model-based gait. These
    `sim-to-real' methods attempt to reduce the gap\footnote{This gap if often referred
    to as the reality gap.} between learning in simulation and real-world evaluation. The work in~\cite{tan2018sim} stresses the importance of realistic
    simulations for sim-to-real policy transfer and develops accurate motor representations with
    simulated noise and latency so that the simulated policies can be used directly on a real robot.
    Rather than carefully modeling the robot dynamics, the work of~\cite{peng2018sim} shows that
    policies trained in simulation with randomized dynamic model parameters (e.g., inertia and friction)
    result in robustness to uncertainty when transferring the policy to a real robot for flat ground locomotion.
    
    Our approach both modulates an existing gait and randomizes parameters to improve the sim-to-real transfer of our policy.
    We also extend dynamics randomization to domain-randomized terrain to train the robot across
    a range of terrains that it may encounter in the real world.

    \textbf{Imitation Learning:}
    Imitation learning uses expert data to inform a learning agent. In
    \cite{peng2020learning}, motion capture data from a dog is adapted for use on a simulated robot using dynamics randomization.
    To transfer the policy to a real robot, advantage-weighted regression is used to maximize the reward on the physical system.
    This method highlights the importance of dynamics randomization and accurate simulation because the reference motions applied
    directly to the robot result in failure. Our approach is similar to imitation learning in that our expert motions
    are given by the open-loop stable Bezier curve gait. However, our policies do not require modification to deploy on a real robot.


\section{Problem Statement} \label{sec:problem_statement}
\newcommand{\Lat}{\rho}
\subsection{General Formulation}

We treat the learning problem as a partially observable Markov
decision process (POMDP) where we have a set of uncertain observations of the
robot's state $o_t \in \mathcal{O}$, a reward function $r_t=\mathcal{R}(s_t,
a_t)$ defining the quality of the locomotion task as a function of
the state $s_t \in \mathcal{S}$, and an action space $a_t \in \mathcal{A}$ containing the set of control inputs to the system. A policy $a_t = \pi(o_t,
\theta)$ maps $o_t$ and some parameters $\theta$ to $a_t$,
which for us includes inputs to a Bezier curve gait generator and
robot foot position residuals.

Given this model, the goal is to find policy parameters $\theta$ such that the
reward is maximized over a finite time horizon $T$ using only partial
observations of the state $o_t$, according to the following objective
\begin{equation}\label{eq:general_rl_problem}
    \theta^\star = \argmax_\theta \mathbb{E}
    \left[
        \sum_{t=0}^{T-1} \gamma^t r_t
    \right],
\end{equation}
where $0 < \gamma \le 1$ is a discount factor, $\theta^\star$ is the optimal
policy parameters, $T$ is the episode length (i.e., how long we simulate the
robot applying $\pi$), and $\mathbb{E}$ is the expected value over the dynamics and domain
($\text{D}^2$) randomized parameters described in Section~\ref{sec:d2_random}.

In simulation, we have access to the full robot state, which we use to construct
the reward function for training; however, full state information is unavailable
on the real robot. Therefore, we train our policy in simulation using a partial
observation of the state $o_t$ that best mimics the sensors on the real-robot.

\subsection{Reinforcement Learning for Gait Modulation} The problem statement
\eqref{eq:general_rl_problem} serves as a template for developing and learning
policies that adapt open-loop gaits for legged locomotion. During each episode,
the policy $\pi$ is applied to modulate and augment a gait generator. This gait
generator (which we describe in more detail in this section) outputs
body-relative foot placements which the robot follows using
inverse-kinematics and joint control.

Let $\ell$ be a label for the quadruped's legs: FL (front-left), FR
(front-right), BL (back-left), BR (back-right). An open loop gait is a
one-dimensional closed parametric curve $\Gamma_\ell(S(t), \zeta, \beta)$
embedded in $\mathbb{R}^3$ and specifying the position of a foot in the frame of
its corresponding hip. Here, $S(t): \mathbb{R} \to [0, 2]$ is a cyclical phase
variable, $\zeta \in \mathbb{R}^{n}$ contains the control inputs and $\beta \in
\mathbb{R}^{m}$ contains the gait parameters (in this paper, $m = 2$). The control inputs $\zeta$ and
gait parameters $\beta$ determine the shape of the gait trajectory. The controls
enable the robot to move in any lateral direction and rotate about its central
axis. Control inputs, parameters, and gaits are discussed
in Section~\ref{sec:bezier}.

The goal of the policy
\begin{equation}
  (\Delta f_{xyz}, \beta) = \pi(o_t, \theta)
\end{equation}
is to augment the foot positions output by the gait functions and to modify the
open-loop gait generator such that (\ref{eq:general_rl_problem}) is maximized
using only partial observations $o_t$. Here $\Delta f_{xyz} =
\begin{bmatrix}\Delta f^{FL}_{xyz}& \Delta f^{FR}_{xyz}& \Delta f^{BL}_{xyz}&
\Delta f^{BR}_{xyz}\end{bmatrix}$, with $\Delta f^\ell_{xyz} \in \mathbb{R}^3$
and $\beta = \{\psi, \delta\}$, where $\psi$ is the clearance height of
the foot above the ground and $\delta$ is a virtual ground penetration depth.

\begin{figure}
\centering
\includegraphics[width=0.5\textwidth]{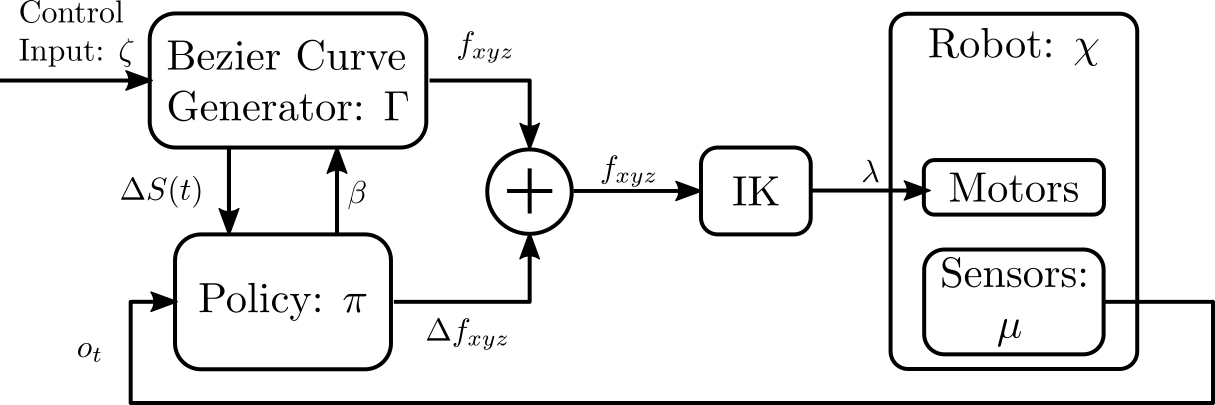}
\caption{D$^2$-GMBC System Diagram}
\label{fig:bez_ctrl}
\end{figure}

The final foot positions are computed as a combination of the gait generator
output and the policy residual:
\begin{equation}\label{gmbc_out}
f_{xyz} = \Gamma (S(t), \zeta, \beta) + \Delta f_{xyz}, \end{equation} where
$f_{xyz} \in \mathbb{R}^{12}$ is the vector of each three-dimensional foot
position that the robot should track and $\Gamma$ contains $\Gamma_\ell$ for
each leg. The policy both adds a residual term to the output and sets the
$\beta$ parameter. Given the foot positions, the robot computes the inverse
kinematics to move its leg joints to the appropriate angles as shown in Fig.~\ref{fig:bez_ctrl}.

The challenge is to solve (\ref{eq:general_rl_problem}) in simulation using only
the observations $o_t$ such that the resulting policy is suitable for use on a
real robot subject to rough terrain and uncertain physical parameters. We
provide a solution to this problem using a combination of a simple policy search
to solve \eqref{eq:general_rl_problem} and dynamics+domain ($\text{D}^2$)
randomization in simulation, which we discuss in Section~\ref{sec:d2_random}. An
overview of the gait modulation is provided in Algorithm~\ref{algo:gmbc} using the
Bezier curve gait generator for $\Gamma$ which we describe further in
Section~\ref{sec:bezier}.


\begin{algorithm}
  \caption{Gait Modulation with Bezier Curves (GMBC)}
  \textbf{Given:} Policy $\pi$ with parameter $\theta$, (Bezier) Curve Generator $\Gamma$,
                    External motion command $\zeta$, robot sensor observations $o_t$, Leg phase $S(t)$
  \begin{algorithmic}[1]
      \State obtain gait modulation from $\pi$ with learned parameter $\theta$
      \State $\Delta f_{xyz}, \beta = \pi (o_t, \theta)$
      \State calculate (Bezier) gait foot placement
      \State $f_{xyz} = \Gamma (S(t), \zeta, \beta)$
    \State \Return $f_{xyz}+ \Delta f_{xyz}$ to robot for IK joint control
  \end{algorithmic}\label{algo:gmbc}
\end{algorithm}

\section{Dynamics + Domain ($\text{D}^2$) Randomization}  \label{sec:d2_random}

We employ two techniques to adapt \eqref{eq:general_rl_problem} for improving
the performance of sim-to-real tranfer of gait modulating policies.
Specifically, this approach merges the ideas from~\cite{peng2018sim} to randomize not only the
dynamics parameters of the simulated robot, but also the terrain that it
traverses. We then solve (\ref{eq:general_rl_problem}) using a simple policy
search method (augmented random search (ARS)~\cite{mania2018simple}) to learn a linear
policy as a function of observations subject to sampled variation in the
dynamics and domain of the simulated episodes \footnote{An episode being a
single $T$ step run of the simulation with a fixed initial (randomly sampled)
state.}. We describe the $\text{D}^2$-randomization below.

\textbf{Dynamics Randomization} We first modify the dynamic
parameters that are known to significantly differ between simulation and reality,
including the mass of each of the robot's links and the friction
between the robot's foot and the ground. The dynamics distribution for
which we train the gait modulating policy is $\sigma_\text{dyn} \sim
\mathbb{P}_\text{dyn}$, where $\sigma$ is the vector of randomized dynamics
parameters which includes the body mass of each link and friction parameter, and
$\mathbb{P}_\text{dyn}$ is a uniform distribution with upper and lower bounds
for the mass of each link and friction parameter (see Table~\ref{tab:d^2} for more
detail). At each training epoch, we sample from $\mathbb{P}_\text{dyn}$ and
run training iteration using these fixed sampled dynamics.

\begin{table}[!htp]\centering
\caption{Dynamics + Domain Randomization}\label{tab:d^2}
\scriptsize
\begin{tabular}{lrr}\toprule
\cellcolor[HTML]{d9d2e9}\textbf{Randomized Parameter $\sigma$} &\cellcolor[HTML]{d9d2e9}\textbf{Range} \\\midrule
Base Mass (Gaussian) &$1.1$kg $\pm 20\%$ \\
Leg Link Masses (Gaussian) &$0.15$kg $\pm 20\%$ \\
Foot Friction (Uniform) &$0.8$ to $1.5$ \\
XYZ Mesh Magnitude (Uniform) &$0$m to $0.08$m \\
\bottomrule
\end{tabular}
\end{table}

\begin{figure}
        \centering
        \includegraphics[width=0.5\textwidth]{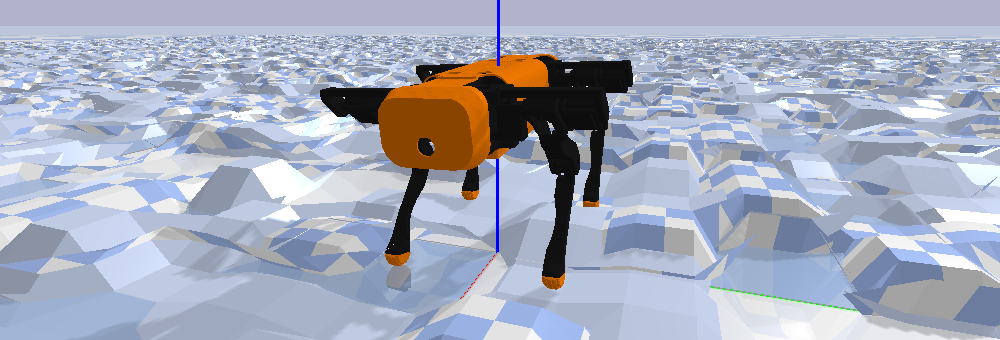}
        \caption{Illustration of the domain randomized terrain used for training in simulation.
            The terrain height varies up to $40\%$ of robot height shown in the image ~\cite{spotminimini2020github}, ~\cite{coumans2017pybullet}.}
        \label{fig:d2rand}
    \end{figure}

\textbf{Domain Randomization} In addition to dynamics randomization, we also perform domain randomization by
randomize the terrain geometry through which the legged robot is moving (see~\ref{fig:d2rand})
We parameterize the terrain as a mesh of
points sampled from $\sigma_\text{dom} \sim \mathbb{P}_\text{dom}$, where
$\sigma_\text{dom}$ is the variation on the uniform mesh grid, and
$\mathbb{P}_\text{dom}$ is a bounded uniform distribution for which we vary the
mesh grid (see Table~\ref{tab:d^2} for more detail). As with dynamics randomization,
we sample the terrain geometry from $\mathbb{P}_\text{dom}$ and train an iteration
of ARS on a fixed sampled terrain.

Let $\sigma \sim \mathbb{P}$ be the joint distribution of
$\mathbb{P}_\text{dyn}$ and $\mathbb{P}_\text{dom}$ over which we take the
expectation in~\eqref{eq:general_rl_problem}, where $\sigma$ is the joint D$^2$
sample. The simulation training is described in more detail in
Algorithm~\ref{algo:d2_training} using ARS and GMBC in
Algorithm~\ref{algo:gmbc}. Policies generated from
Algorithm\ref{algo:d2_training} are defined as D$^2$-GMBC policies. During
training, the external commands $\zeta$ are held fixed and defined by the task
i.e., move forward at a fixed velocity.

\begin{algorithm}
  \caption{RL Simulation training for D$^2$-Randomized GMBC (D$^2$-GMBC) using Augmented Random Search~\cite{mania2018simple}}
  \textbf{Initialize:} policy parameters $\theta_0$, D$^2$ distribution $\mathbb{P}$, reward function $\mathcal{R}$,
            GMBC (Algorithm~\ref{algo:gmbc}), iteration number $k=0$, construct ARS.
  \begin{algorithmic}[1]
      \While {training not converged}
            \State $\sigma \sim \mathbb{P}$ sample D$^2$ parameters
            \State ARS step of (\ref{eq:general_rl_problem}) with D$^2$-randomization + GMBC
            \State $\theta_{k+1} \gets \text{ARS}(\pi, \theta_k, \mathcal{R}, \sigma, k)$
            \State $k \gets k + 1$
      \EndWhile
      \State \Return $\theta_k$
  \end{algorithmic}\label{algo:d2_training}
\end{algorithm}

In the next section, we describe the specific Bezier curve gait
generator $\Gamma$ used throughout this work.


\section{Extending Bezier Curves using Motion Commands}\label{sec:bezier}

D$^2$-GMBC uses an extended and open-loop version of the Bezier curve gaits developed in~\cite{cheetah2014}, combining
multiple 2D gaits into a single 3D gait that enables transverse, lateral, and rotational motion.
Our policy constantly modifies these trajectories, adapting them to uneven terrain while removing the need to sense foot impacts.

A gait trajectory is a closed curve that a foot follows during locomotion. It consists of two phases: swing and stance. During swing, the foot moves
through the air to its next position. During stance, the foot contacts the ground and moves the robot using ground reaction forces.
A gait is parameterized by a phase $S(t) \in [0,2)$, which determines the foot's location along the trajectory. The leg is in stance for $S(t) \in [0, 1)$ and
in swing for $S(t) \in [1, 2)$.

A trajectory generator $\Upsilon(S(t), \tau)$ maps phase and step length $\tau$ to a set of trajectories in $\mathbb{R}^2$.
We use a trajectory consisting of a Bezier curve during swing and a sinusoidal curve during stance, see Section~\ref{subsec:2dbezier} for details.

To robot has three control inputs: $\zeta = \begin{bmatrix}\rho & \bar{\omega} & L_\text{span}\end{bmatrix}$, where $\rho \in [-\frac{\pi}{2}, \frac{\pi}{2}]$
is the trajectory's rotation angle relative to the robot's forward direction, $\bar{\omega}$ is the robot's yaw velocity, and $L_\text{span}$ is half the stride length.

\begin{figure*}
          \centering
          \includegraphics[width=0.45\textwidth]{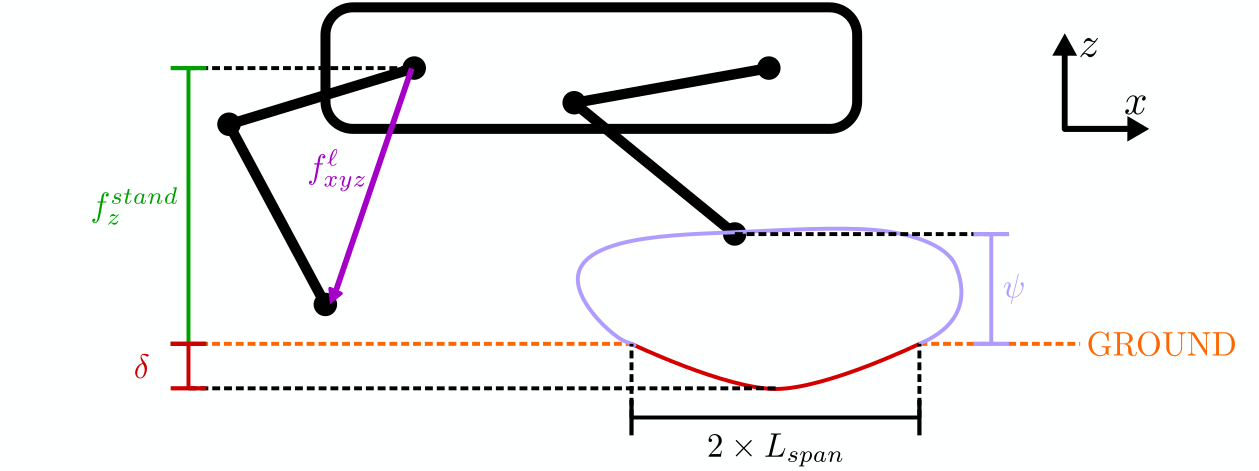}
          \includegraphics[width=0.45\textwidth]{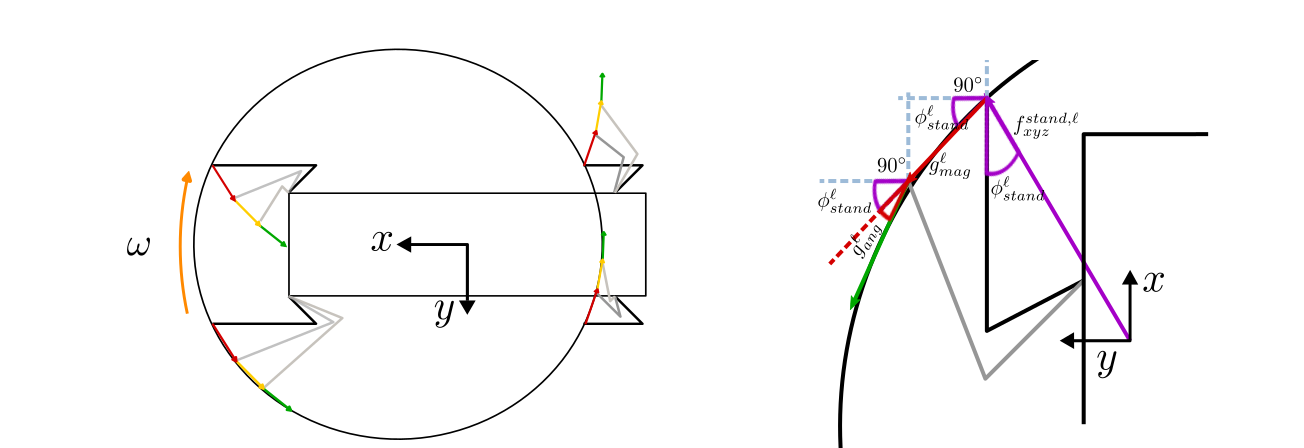}
          \caption{Schematic of foot placement based on Bezier Gait Generator: $f_{xyz}$.}
          \label{ikbezier}
        \end{figure*}

Locomotion consists of a planar translation $f^{tr}_{qz} = \Upsilon(S(t), L_\text{span})$ and yaw trajectory
$f^{yaw}_{qz} = \Upsilon(S(t), \bar{\omega})$. These trajectories also depend on curve parameters $\beta = \begin{bmatrix}\psi & \delta \end{bmatrix}$ (see Fig.~\ref{ikbezier} and Section~\ref{subsec:2dbezier}).

  Using the control inputs, we convert the planar trajectories $f^{tr}_{qz}$ and $f^{yaw}_{qz}$ into 3D foot-position trajectories $f_{xyz}^{tr}$
  and $f_{xyz}^{yaw, \ell}$, where each leg $\ell$ has the same translational velocity but different yaw velocity.
  Finally, we transform the yaw and translational curves into a frame relative to each leg's rest position $f_{xyz}^\mathrm{stand}$ to get the final foot trajectory
  for leg $\ell$:
  
\begin{equation}\label{eq:fxyz}
    f^{\ell}_{xyz} = f_{xyz}^{tr} + f_{xyz,\ell}^{yaw} + f_{xyz}^\mathrm{stand}.
\end{equation}

This scheme enables movements encompassing forward, lateral, and yaw commands and enables policies for straight-line motion to
extend to more complex motions.

In particular, let $(f^{tr}_{q}, f^{tr}_{z})$ and $(f^{yaw}_{q}, f^{yaw}_z)$ be the coordinates of $f^{tr}_{qz} \in \mathbb{R}^2$ and $f^{yaw}_{qz} \in \mathbb{R}^2$ respectively.
Rotating planar trajectory $f_{qz}^{tr}$ by $\Lat$ yields the 3D foot trajectory:
\begin{equation}\label{eq:tr}
  f_{xyz}^{tr} =\begin{bmatrix} f_{q}^{tr}\cos{\Lat} & f_{q}^{tr} \sin{\Lat} & f_{z}^{tr}\end{bmatrix}.
\end{equation}

Yaw control is inspired by a four-wheel steered car~\cite{lee2014turning}. 
To trace a circular path, each foot path's angle must remain tangent
to the rotational circle, as shown in Fig.~\ref{ikbezier}.

The yaw command for foot $\ell$, $g_{xzy}^\ell(t)$, depends on the previous foot position relative to $f_{xyz}^\text{stand}$ (with $g_{xyz}^\ell(0) = 0$):
\begin{equation}
  g_{xyz}^\ell(t) = f_{xyz}(t-1) - f_{xyz}^\text{stand}.
\end{equation}

The distance and angle of this step in the $xy$ plane are:
\begin{align}
  g_\text{mag}^\ell &= \sqrt{(g_{x}^\ell)^2 + (g_{y}^\ell)^2}, \\
  g_\text{ang}^\ell &= \arctan{\frac{g_{y}^\ell}{g_{x}^\ell}},
\end{align}
where $g_{x}^\ell$ and $g_{y}^\ell$ are the $x$ and $y$ components of $g_{xyz}^\ell$.

Each leg translates at an angle for a given yaw motion:
\begin{equation} \label{phi_arc}
  \phi^{\ell}_\text{arc} = g_\text{ang}^\ell + \phi_\text{stand}^\ell + \frac{\pi}{2},
\end{equation}

\begin{equation} \label{phi_arc}
  \phi_\text{stand}^\ell = \begin{cases}
        \arctan{\frac{f_{y}^\text{stand}}{f_{x}^\text{stand}}} & \ell = \text{FR, BL}\\
        -\arctan{\frac{f_{y}^\text{stand}}{f_{x}^\text{stand}}} & \ell = \text{FL, BR}
  
  \end{cases}
\end{equation}
where $f_{x}^\text{stand}$ and $f_{y}^\text{stand}$ are the $x$ and $y$ components
of $f_{xyz}^\text{stand}$. The leg rotation angle $\phi^\ell_{arc}$ provides the final yaw trajectory:
\begin{equation}\label{eq:yaw}
  f_{xyz,\ell}^{yaw} = \begin{bmatrix} f_{q}^{yaw}\cos{\phi^\ell_{arc}} &f_{q}\sin{\phi^\ell_{arc}} & f_{z}^{yaw} \end{bmatrix}
\end{equation}

Algorithm \ref{algo:gait_ctrl} describes the entire gait generation process, and Fig.~\ref{fig:bez_ctrl} shows the Bezier controller system diagram.

\begin{algorithm}
  \caption{Bezier Curve Generator $\Gamma$ - Per Leg $\ell$}\label{algo:gait_ctrl}
  \textbf{Inputs:} $\zeta$
  \begin{algorithmic}[1]
    \State Map $t$ to foot phase $S_{\ell}(t)$ using \eqref{S_stance}
    \State $(f_{q}^{tr}, f_{z}^{tr}) = \Upsilon(S_\ell(t), L_\mathrm{span})$
    \State $(f_{q}^{yaw}, f_{z}^{yaw}) = \Upsilon(S_\ell(t), \bar{\omega})$
    \State $f_{xyz}^{tr} =\begin{bmatrix} f_{q}^{tr}\cos{\Lat} & f_{q}^{tr} \sin{\Lat} & f_{z}^{tr} \end{bmatrix}$
    \State $\phi^{\ell}_\text{arc} = g_\text{ang}^\ell + \phi_\text{stand}^\ell + \frac{\pi}{2}$
    \State $f_{xyz,\ell}^{yaw} = \begin{bmatrix} f_{q}^{yaw}\cos{\phi^\ell_{arc}} &f_{q}\sin{\phi^\ell_{arc}} & f_{z}^{yaw} \end{bmatrix}$
    \State $f^{\ell}_{xyz} = f_{xyz}^{tr} + f_{xyz,\ell}^{yaw} + f_{xyz}^\mathrm{stand}$
    \State \Return $f^{\ell}_{xyz}$ to the robot for joint actuation
  \end{algorithmic}
\end{algorithm}

\section{Results} \label{sec:sim_results}

We now present several experiments to evaluate our approach,
D$^2$-GMBC, in generating legged locomotion for sim-to-real transfer. We first
describe the simulated training in more detail, explaining the
robot observations, the policy structure, and the reward function.
We then to evaluate D$^2$-GMBC on the following criteria:
\begin{enumerate}
    \item Generalization to randomized dynamics and terrain.
    \item Improvement over open-loop gait generators with and without D$^2$-Randomization.
    \item Sim-to-real transfer performance on a real robot.
\end{enumerate}

\subsection{Simulated Training}
    We first explitly describe the simulated training of D$^2$-GMBC. As
    mentioned in Algorithm~\ref{algo:d2_training}, we use the augmented random
    search (ARS) method to train a policy to modulate GMBC
    (Algorithm~\ref{algo:gmbc}) using the objective function defined
    in~\eqref{eq:general_rl_problem}. To match the sensors on the real-robot,
    we train the policy using an observation comprised of
    body roll $r$ and pitch $p$ angles relative to the gravity vector, the body
    3-axis angular velocity $\omega$, the body 3-axis linear acceleration
    $\dot{v}$, and the interal phase of each foot $S_\ell(t)$, making $o_t = [r,
    p, \omega, \dot{v}, S(t) ]^\top \in \mathbb{R}^{12}$.

    We chose a linear policy so that it could run in real-time on inexpensive hardware while improving the open-loop gaits as best as possible.
    The policy is
    \begin{equation*}
        a_t  = \pi(o_t,\theta) = \theta^\top o_t,
    \end{equation*}
    where $\theta \in \mathbb{R}^{12 \times 14}$. Here, the policy outputs the
    nominal clearance height and virtual ground penetration depth of
    the Bezier curve and residual foot displacements that are added to the
    output of the Bezier curve.

\begin{figure}
    \centering
    \includegraphics[width=0.5\textwidth]{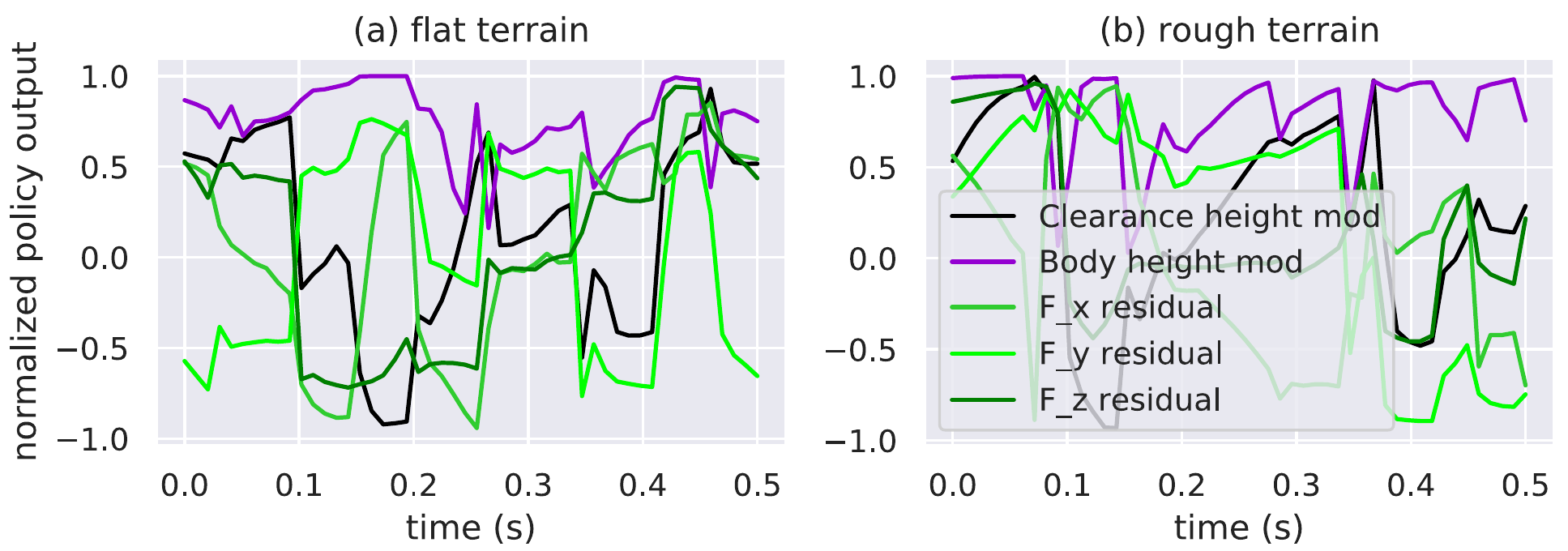}
    \caption{Example policy output during testing for a single leg. The simple linear policy
        successfully modulates robot gaits using only on-board inertial measurements.
    }\label{fig:policy_out}
    \end{figure}

    To prevent infeasible foot positions and Bezier curve parameters,
    we center the policy output within the domain $\pi(o_t, \theta) \in [-1, 1]^{14}$.
    The output is then remapped to the acceptable range of Bezier curve parameters and a bounded
    domain of foot residuals. For each of the simulated examples, the high-level
    motion commands are fixed at $L_{span} = 0.035$m, and $\Lat = 0$.
    A proportional controller sets the yaw rate $\bar{\omega}$ to ensure that the robot's heading is always zero.
    Fig.~\ref{fig:policy_out} provides an example of a D$^2$-GMBC policy for a single leg.

    Augmented random search uses a random parameter search policy
    optimization technique for reinforcement learning. Thus, for each ARS
    optimization step, we deploy 16 rollouts per iteration with a parameter
    learning rate of 0.03 and parameter exploration noise of 0.05. That is, each
    of the 16 rollouts samples a new parameter $\bar{\theta} = \theta + \Delta
    \theta$ where $\Delta \theta \sim \mathcal{N}(\mathbf{0}, 0.05)$ where
    $\mathcal{N}$ is a normal distribution with mean $\mathbf{0} \in
    \mathbb{R}^{12 \times 14}$ and variance $0.05$. Each episode lasts $T=5000$
    steps (50 seconds). The reward function is
    \begin{equation}
        r_t = \Delta x -10 \left(|r| + |p| \right) - 0.03\sum|\omega|
    \end{equation}
    where $\Delta x$ is the global distance traveled by the robot in the
    horizontal $x$-direction in one time step.
    We found that dividing the final episode reward by the number
    of time steps improves the policy's learning
    because it reduces the penalty for a sudden fall after an otherwise successful run, and ultimately encourages survivability. For D$^2$-randomization
    training, we resample a new set of D$^2$ parameters (see Sec.~\ref{sec:d2_random}) at each training episode of ARS.
\subsection{Improvement to GMBC}

    To compare the effectiveness of randomizing the robot's dynamics and
    the terrain for legged locomotion, we train a
    GMBC policy in simulation with and without D$^2$-randomization and benchmark
    the policies on unforseen terrain and dynamics. We then measure how far the robot can travel using each method before it fails. We compare the results
    of GMBC and D$^2$-GMBC to the open-loop Bezier curve gait generator
    (unaugmented) by counting the number
    of times the robot did not fall (exceeded a roll or pitch of $60^{\circ}$ or hit
    the ground) within the allotted simulation time of 50,000 steps, or 500 seconds. We evaluate this
    over a set of 1000 trials with randomized dynamics and terrain.

\begin{figure}
          \centering
          \includegraphics[width=0.35\textwidth]{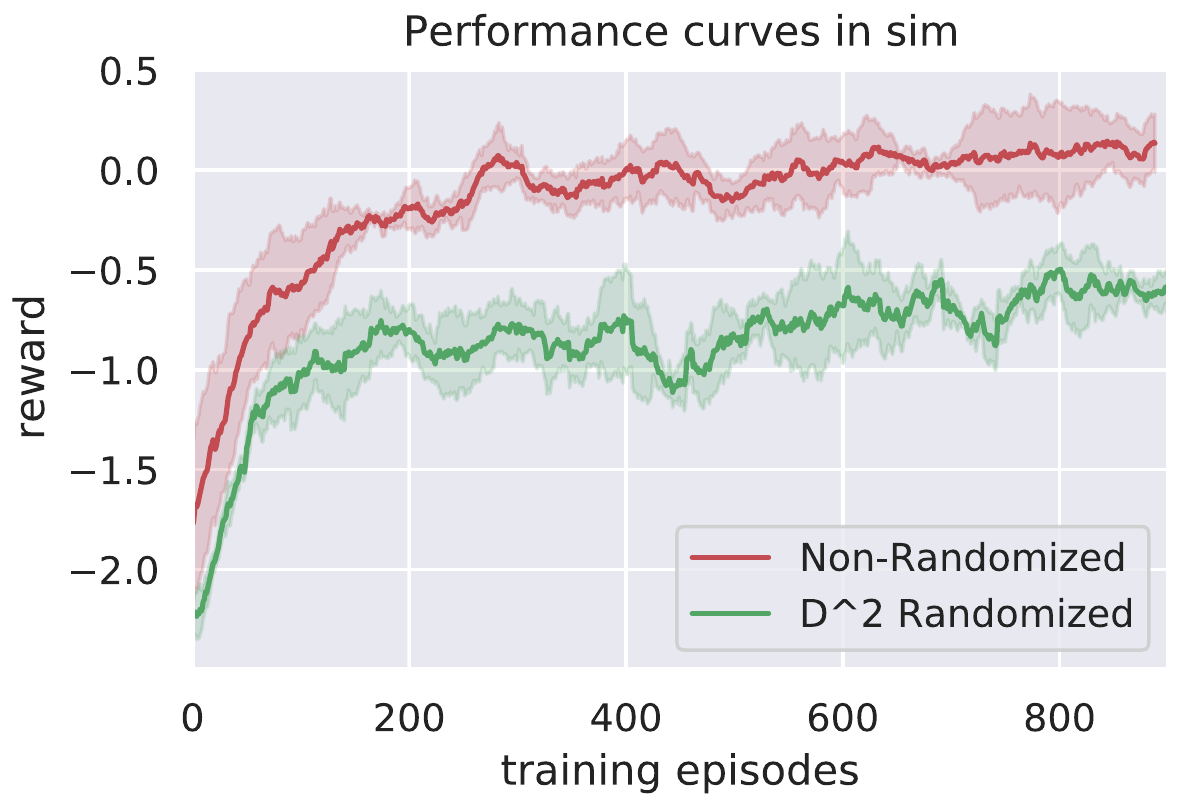}
          \caption{
                  Evaluated reward curves for simulation training with and
                  without $\text{D}^2$ randomization. Note that despite the
                  improved performance with non-randomization, policies trained
                  with $\text{D}^2$ randomization transfer better to unforseen
                  dynamics and terrain as shown in Table~\ref{tab:simexp}.
          }
          \label{fig:sim_training}
        \end{figure}

    As shown in Table.~\ref{tab:simexp}, robots using D$^2$-GMBC have improved survivability on randomized dynamics and terrain when compared
    to GMBC policies trained on a single set of dynamics and terrain environment
    parameters; out of 1000 trials, 146 out of 305 (45\%) D$^2$-GMBC survivals traveled past 90m, and only 26 out of 327 (8\%) GMBC survivals did the same, showing a $5.6 \times$ improvement. Notably, both vastly outperformed the open-loop gait trials, which never made it past 5m and did not sruvive a single run. The training curves for the D$^2$-GMBC and
    GMBC (non-randomized) policies in Fig.~\ref{fig:sim_training} depict that
    the non-randomized GMBC policy is expected to perform significantly better
    than D$^2$-GMBC policies. These data highlight the potential for
    simple linear policies to perform well in legged locomotion tasks and provide evidence that
    training performance is not a useful indicator for testing policy-based
    locomotion skills in sim-to-real settings.

    \begin{table}[!htp]\centering
      \caption{Each method was tested for 1000 trials, each lasting 50,000 timesteps.  The distribution of maximum distances traveled by the agent
        and whether it fell or lived are reported.}\label{tab:simexp}
    \scriptsize
    \begin{tabular}{lrr|rr|rr}\toprule
     &\multicolumn{2}{c}{\cellcolor[HTML]{d9d2e9}\textbf{D$^2$-GMBC}} &\multicolumn{2}{c}{\cellcolor[HTML]{d9d2e9}\textbf{GMBC}} &\multicolumn{2}{c}{\cellcolor[HTML]{d9d2e9}\textbf{Open-loop}} \\\cmidrule{1-7}
    \cellcolor[HTML]{d9d2e9}\textbf{Distance} &\cellcolor[HTML]{d9d2e9}\textbf{\# Died} &\cellcolor[HTML]{d9d2e9}\textbf{\# Lived} &\cellcolor[HTML]{d9d2e9}\textbf{\# Died} &\cellcolor[HTML]{d9d2e9}\textbf{\# Lived} &\cellcolor[HTML]{d9d2e9}\textbf{\# Died} &\cellcolor[HTML]{d9d2e9}\textbf{\# Lived}\\\midrule
    \cellcolor[HTML]{d9d2e9}\textbf{$\leq$ 5m} &488 &64 &450 &121 &1000 &0 \\
    \cellcolor[HTML]{d9d2e9}\textbf{5m to 90m} &207 &95 &222 &180 &N/A &N/A \\
    \cellcolor[HTML]{d9d2e9}\textbf{$\geq$ 90m} &0 &146 &1 &26 &N/A &N/A \\
    \bottomrule
    \end{tabular}
\end{table}

\subsection{Sim-to-Real Transfer} \label{sec:real_results}
    The previous simulations illustrate the generalization capabilies of
    the linear policy for GMBC using D$^2$ randomization:
    training with D$^2$-randomization improves legged locomotion performance.
    We now describe three experiments to demonstrate the sim-to-real transfer capabilities of D$^2$-GMBC. We conduct three experiments on OpenQuadruped ~\cite{openquadruped2020}, an inexpensive open-source robot built with hobbyist components.

\begin{figure}
      \centering
      \includegraphics[width=0.45\textwidth]{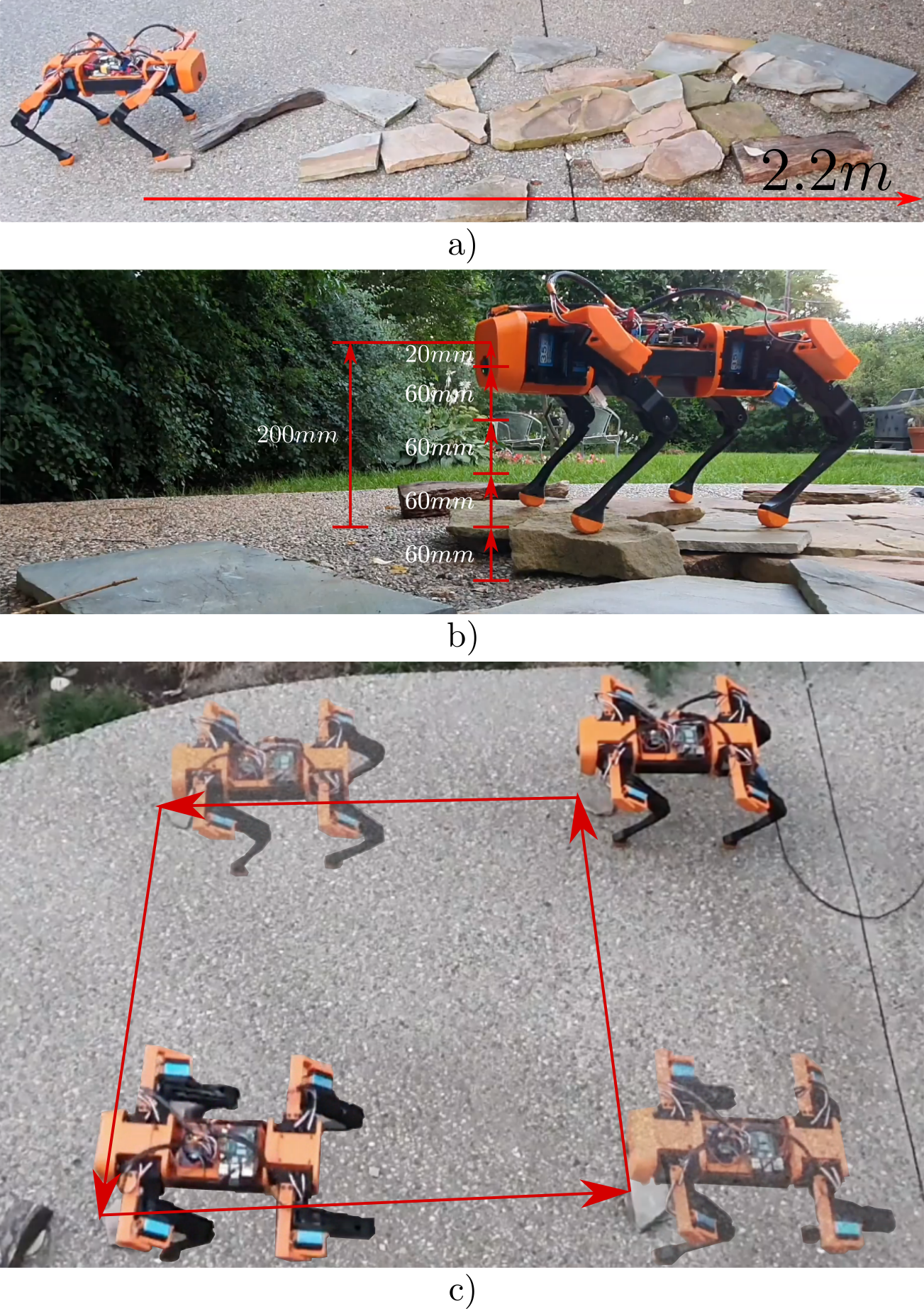}
      \caption{Illustration of experimental testbeds: a) Experiment 1: Rocky Test Track (2.2m), b) Experiment 2: $60mm$ Loose Stone Descent, c) Experiment 3: Omnidirectional Performance on Flat Ground. Media available at \url{https://sites.google.com/view/d2gmbc/home}.}
      \label{fig:experiments}
    \end{figure}

    The first experiment involves testing D$^2$-GMBC on a robot whose task is to
    traverse the $2.2$ m track shown in Fig.~\ref{fig:experiments} (a) covered
    with loose stones whose heights range between $10$ mm to $60$ mm (roughly
    $30\%$ of the robot's standing height). The second test evaluates the
    robot's ability to descend from the peak of loose stones at the maximum $60$
    mm height onto flat ground shown in Fig.~\ref{fig:experiments} (b). The final test, shows
    the generalization capabilies of D$^2$-GMBC trained in simulation for following
    unseen high-level motion commands $\zeta$ by having the robot follow
    the $1\times1$ m square shown in Fig.~\ref{fig:experiments}(c).

    \textbf{Experiment 1: Traversing Unknown Terrain} In this experiment, we
    test the policy on terrain that was not seen in training. Additionally,
    the stones are loose, making the terrain non-stationary and potentially difficult to traverse.
    Due to the lack of a global odometry, a human operator provided high-level yaw rates to rectify
    the robot traversing the stones. To prevent human bias, the robot randomly selected with a 50\% probability
    the D$^2$-GMBC gait or a benchmark open-loop Bezier curve gait. The experiment continued until
    both methods were used for at least 10 trials.

    \begin{table}[!htp]\centering
    \caption{Experiment 1: Rocky Test Track (2.2m).}\label{tab:exp1}
    \scriptsize
    \begin{tabular}{lrrr}\toprule
    &\cellcolor[HTML]{d9d2e9}\textbf{D$^2$-GMBC} &\cellcolor[HTML]{d9d2e9}\textbf{Open-loop Bezier Gait} \\\midrule
    \cellcolor[HTML]{d9d2e9}\textbf{Distance Mean (of 2.2)} &1.93 &1.37 \\
    \cellcolor[HTML]{d9d2e9}\textbf{Std. Dev} &0.30 &0.44 \\
    \cellcolor[HTML]{d9d2e9}\textbf{Success Rate (of 1)} &0.60 &0.14 \\
    \cellcolor[HTML]{d9d2e9}\textbf{Distance Improvement} &40.73\% & \\
    \cellcolor[HTML]{d9d2e9}\textbf{Success Improvement} &4.28 $\times$ & \\
    \bottomrule
    \end{tabular}
        \end{table}

    As shown in Table~\ref{tab:exp1}, we observed a $1.408 \times$ increase in
    average traversed distance using D$^2$-GMBC compared to the open-loop gait.
    Although the track was only $2.2$ m long, D$^2$-GMBC improved the survivability
    by $4.28 \times $ compared to the open-loop Bezier gait.
    Furthermore, our approach does not require sensing leg joint positions or foot contact.

    \textbf{Experiment 2: Descending from Loose Stones} Here, we
    set the robot's starting position on a raised platform consisting of
    $60mm$ stones to perform a descent test. We record
    the ratio between successful and failed descents for both D$^2$-GMBC and open-loop controllers.
    The operator, who does not know which policy is being used, drives the robot forward until it either
    descends or falls. The D$^2$-GMBC agent fell $3$ out of $11$ times and the open-loop controller fell $9$
    out of $13$ times, showing a $2.36 \times$ improvement rate on a test with a sharp descent and without contact sensing.

    \textbf{Experiment 3: External Command Generalization} This
    experiment runs the robot on flat terrain to validate the generalization of D$^2$-GMBC to external yaw and lateral
    commands after the sim-to-real transfer and to provide evidence that the improved robustness to rough
    terrain does not negatively affect performance on flat terrain, even though the policy was trained exclusively for
    rough terrain.

    The operator drove the robot around a $1m \times 1m$ track, performing
    forward, backward and strafing motions, with some yaw commands to correct
    the robot's heading if necessary. Aside from showing the versatility of
    D$^2$-R GMBC in seamlessly providing the operator with mixed motion control,
    the test allowed us to measure locomotion speed by correlating video timestamps with
    marks on the ground.

    In our tests, the D$^2$GMBC policy, compared to the open-loop gait, was $11.5 \%$ faster strafing to the left,
    $19.1 \%$ faster strafing to the right, and had the same  forward speed.  Backward speed, however, fell by $57.6 \%$.
    Further inspection revealed that this outlier result was due to the walking stance used to increase robustness
    causing the two rear motors (both 23kg hobby servos) to reach their torque limits and the rear of the robot to dip. The policy then damped the
    robot's motion in anticipation of a fall. Simulation evidence indicates
    that with more powerful motors, backwards walking would experience no performance degradation.

\begin{table}[!htp]\centering
\caption{Experiment 3: Omnidirectional Speed on Flat Ground.}\label{tab:exp2}
\scriptsize
\begin{tabular}{lrrrrr}\toprule
\cellcolor[HTML]{d9d2e9}\textbf{Gait} &\cellcolor[HTML]{d9d2e9}\textbf{FWD (m/s)} &\cellcolor[HTML]{d9d2e9}\textbf{LEFT (m/s)} &\cellcolor[HTML]{d9d2e9}\textbf{BWD (m/s)} &\cellcolor[HTML]{d9d2e9}\textbf{RIGHT (m/s)} \\\midrule
D$^2$-GMBC AVG &0.21 &0.29 &0.15 &0.25 \\
D$^2$-GMBC STD &0.04 &0.04 &0.03 &0.05 \\
Open-Loop AVG &0.20 &0.26 &0.26 &0.21 \\
Open-Loop STD &0.02 &0.04 &0.09 &0.04 \\
\bottomrule
\end{tabular}
\end{table}

\section{Conclusion} \label{sec:conclusion}
We present a new control architecture (D$^2$-GMBC) that augments an open-loop stable gait with a learned policy.
Formulating the problem as a POMDP using only IMU data to inform our policy, we generate stable locomotion on unobserved rough terrain.
With D$^2$-GMBC, policies trained exclusively in simulation with D$^2$ randomization can be directly deployed on real-robots.
The method is data efficient, achieving a $5.6 \times$ high-distance ($\geq$90m) survival on randomized terrain relative to
a similar learning procedure without D$^2$ randomization in fewer than 600 training epochs. Real robot tests show $4.28 \times$ higher survivability and significantly
farther travel than a robot using an open-loop gait, despite the terrain being vastly different from the simulated mesh, including having the dynamics of loose stones. On flat ground, omnidirectional speed is preserved.
For future work, we are interested to see how this method can be carried over to torque-controllable systems to achieve more dynamic behaviors without terrain sensing.

\section*{APPENDIX}
\subsection{2D Bezier Curve Gait} \label{subsec:2dbezier}

We discuss the Bezier curve trajectories developed in~\cite{cheetah2014}.

\subsubsection{Trajectory Generation} \label{subsubsec:bezier}

Each foot's trajectory is
\begin{equation}\label{tempgait}
  \Upsilon(S(t), \tau) = \begin{cases}
  \begin{bmatrix}\tau(1 - 2 S(t)) \\ \delta \cos{\frac{\pi \tau(1 - 2 S(t))}{2\tau}}\end{bmatrix} & 0 \leq S(t) < 1,\\
  \sum\limits_{k=0}^n c_k(\tau, \psi) B^n_k(S(t) - 1) & 1 \leq S(t) < 2 \\
    \end{cases},
\end{equation}
a closed parametric curve with
\begin{equation}
  B^n_k(S(t)) = \binom{n}{k}(1 - S(t))^{(n - k)}S(t).
\end{equation}
Here $B^n_k(S(t))$ is the Bernstein polynomial~\cite{lorentz1986bernstein} of degree $n$ with $n+1$ control points $c_k(\tau, \psi) \in \mathbb{R}^2$.
Our Bezier curves use 12 control points (see Table~\ref{ctrl_pts}).  The parameter $\tau$ determines the curve's shape and
$0 \leq S(t) \leq 2$ determines the position along the curve. The stance phase is when $0 \leq S(t) \leq 1$ and the swing phase is when $1 \leq S(t) < 2$.

\begin{table}[!htp]\centering
\caption{Bezier Curve Control Points. ~\cite{cheetah2014}} \label{ctrl_pts}
\scriptsize
\begin{tabular}{lrrrrr}\toprule
\cellcolor[HTML]{d9d2e9}\textbf{Control Point} &\cellcolor[HTML]{d9d2e9}(\textbf{q}, \textbf{z}) & \cellcolor[HTML]{d9d2e9}\textbf{Control Point} &\cellcolor[HTML]{d9d2e9}(\textbf{q}, \textbf{z}) \\\midrule
$c_0$          & $(-\tau,0.0 )$ &$c_7$      & $(0.0, 1.1 \psi)$\\
$c_1$          & $(-1.4 \tau,0.0)$ &$c_8, c_9$ &$(-1.5 \tau, 1.1 \psi)$  \\
$c_2, c_3, c_4$ & $(-1.5 \tau ,0.9 \psi)$& $c_{10}$  & $(-1.4 \tau ,0.0)$\\
$c_5, c_6$      & $(0.0,0.9 \psi )$& $c_{11}$   & $(\tau,0.0)$   \\
\bottomrule
\end{tabular}
\end{table}

\subsubsection{Leg Phases} \label{subsubsec:phase}
During locomotion, each foot follows a periodic gait trajectory.
The total time for the legs to complete a gait cycle is $T_\text{stride} = T_\text{swing} + T_\text{stance}$, where $T_\text{swing}$ is the duration
of the swing phase and $T_\text{stance}$ is the duration of the stance phase.  We determine $T_\text{swing}$ empirically (0.2 seconds in our case)
and set $T_{stance} = \frac{2 L_\text{span}}{v_d}$, where $v_d$ is a fixed step velocity and $L_\text{span}$ is half of the stride length.

The relative timing between the swing and stance phases of each leg determines which legs are touching the ground and which swing
freely at any given time. Different phase lags between each leg correspond to different qualitative locomotion types such as walking, trotting, and galloping.
This periodic motion lets us determine the position of every leg relative to $t^\text{elapsed}_\mathrm{FL}$, the most recent time the
front-left leg has impacted the ground. Since our robot does not sense contacts, we reset $t^\text{elapsed}_\mathrm{FL}$ to 0 every $T_{stride}$ seconds.

We define the clock for each leg $\ell$ to be
\begin{equation}
  t_\ell = t^{elapse}_\mathrm{FL} - \Delta S_{\ell}(t) T_{stride},
\end{equation}
where $\Delta S_\ell$ is the phase lag between the front-left leg and $\ell$.

\begin{figure}
          \centering
          \includegraphics[width=0.45\textwidth]{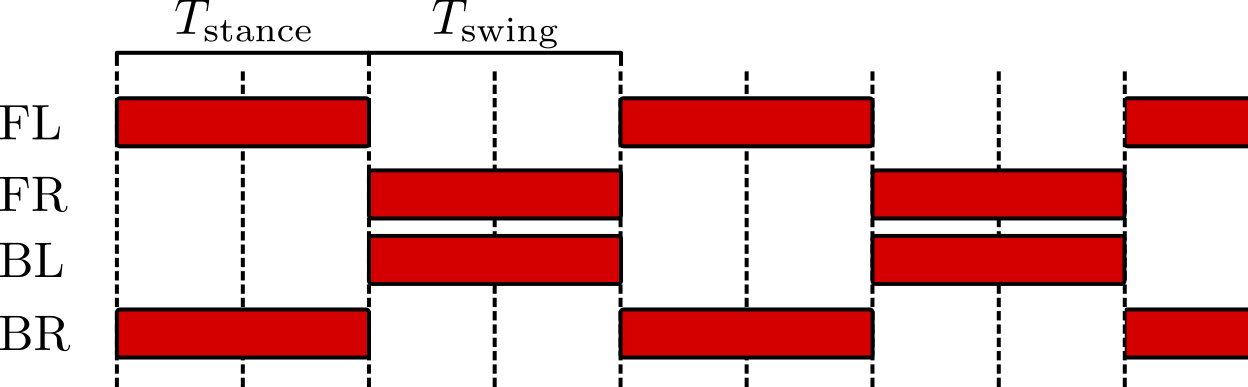}
          \caption{Leg phases for trotting, where Red is Stance. $T_\text{stance} = T_\text{swing}$}
          \label{legphases}
        \end{figure}

We set the relative phase lag $\Delta S_\ell$ to create a trotting gait:
\begin{equation}\label{phaselag}
   \begin{bmatrix}
    \Delta S_{FL} \\
    \Delta S_{FR} \\
    \Delta S_{BL} \\
    \Delta S_{BR}
  \end{bmatrix} =
  \begin{bmatrix}
    0.0 \\
    0.5 \\
    0.5 \\
    0.0
  \end{bmatrix}
\end{equation}
The relative timing of the swing and stance phases for such a gait are depicted in Fig.~\ref{legphases}.

Next, we normalize the clock for each leg, mapping $t_\ell$ to the parameter $S_\ell(t)$ such that
the leg in stance when $0 \leq S_\ell(t)  < 1$ and in swing when $1 \leq S_\ell(t) \leq 2$:
\begin{equation}\label{S_stance}
        S_\ell(t) = \begin{cases}
                        \frac{t_\ell}{T_{stance}} & 0 < t_\ell < T_{stance}\\
                        \frac{t_\ell + T_{stride}}{T_{stance}} &  -T_{stride} < t_\ell < -T_{swing} \\
                        \frac{t_\ell + T_{swing}}{T_{swing}} &  -T_{swing} < t_\ell < 0 \\
                        \frac{t_\ell - T_{stance}}{T_{swing}} &  T_{stance} < t_\ell < T_{stride}
                    \end{cases}
                  \end{equation}
For the first two cases in~\eqref{S_stance} the legs are in stance and the last two cases they are in swing.

We can then use $S_\ell(t)$ to compute the foot position for leg $\ell$ using~\eqref{tempgait}. Such a scheme would suffice for forward walking.
However, because we need to walk laterally and turn, we employ the techniques used in section ~\ref{sec:bezier}.


\section*{ACKNOWLEDGMENT}

Thank you to Adham Elarabawy for co-creating and collaborating on the development of OpenQuadruped.

\bibliographystyle{IEEEtran.bst}
\bibliography{main.bib}  

\end{document}